\documentclass[a4paper]{article}

\usepackage{INTERSPEECH2020}
 \usepackage{booktabs}
 \usepackage{multirow}
 \usepackage{multicol}
 \usepackage{tabu}
 \usepackage{xcolor,colortbl}
 \usepackage{xcolor}
\newcolumntype{?}{!{\vrule width 1.1pt}}
\newcommand{\hbl}{\noalign{
\hrule height 1.1pt
}}
\newcommand{\pad}{1.3}

\begin{document}

\title{Cross-lingual and Multilingual Spoken Term Detection for\\ Low-Resource Indian Languages}
 %
 \name{Sanket Shah\textsuperscript{$+$}, Satarupa Guha\textsuperscript{$*$}, Simran Khanuja\textsuperscript{$+$}, Sunayana Sitaram\textsuperscript{$+$}}
 \address{\textsuperscript{$+$}Microsoft Research India, Bangalore, India \\\textsuperscript{$*$}Microsoft Corporation, Hyderabad, India}
 \email{\{t-sansha, saguha, t-sikha, sunayana.sitaram\}@microsoft.com}
 %

 %
 \maketitle
 \begin{abstract}

Spoken Term Detection (STD) is the task of searching for words or phrases within audio, given either text or spoken input as a query. In this work, we use state-of-the-art Hindi, Tamil and Telugu ASR systems cross-lingually for lexical Spoken Term Detection in ten low-resource Indian languages. Since no publicly available dataset exists for Spoken Term Detection in these languages, we create a new dataset using a publicly available TTS dataset. We report a standard metric for STD, Mean Term Weighted Value (MTWV) and show that ASR systems built in languages that are phonetically similar to the target languages have higher accuracy, however, it is also possible to get high MTWV scores for dissimilar languages by using a relaxed phone matching algorithm. We propose a technique to bootstrap the Grapheme-to-Phoneme (g2p) mapping between all the languages under consideration using publicly available resources. Gains are obtained when we combine the output of multiple ASR systems and when we use language-specific Language Models. We show that it is possible to perform STD cross-lingually in a zero-shot manner without the need for any language-specific speech data. We plan to make the STD dataset available for other researchers interested in cross-lingual STD.

 \end{abstract}
 \section{Introduction}
 \label{sec:intro}

Spoken Term Detection (STD) or Keyword Spotting is the problem of finding a word or phrase in audio and is used for various applications such as audio search and indexing. In this work, we focus on lexical STD, in which the target term is specified in the text form and needs to be found in audio. A related problem that we do not address is that of Query-by-Example (QbE) STD in which the term to be detected is specified in spoken form.

A key component of an STD system is an Automatic Speech Recognition system which can be run on the audio to produce a transcription. If the ASR system is accurate, the task of lexical STD is trivial, as the term we are looking for can be searched for in the ASR transcript, and timestamp information can be recovered. However, in low resource languages, either an ASR system does not exist, or is not accurate enough, and enough data to train an ASR system may not be available.

Although there have been efforts in building zero-shot ASR systems in the past, ASR systems still have low accuracies if the target language has no speech and text data for building Acoustic and Language Models. A Spoken Term Detection system typically only searches for specific keywords in the audio. Intuitively, STD systems should be easier to build in a zero-shot manner cross-lingually than a full-fledged ASR system. An STD system can be extremely useful for users who can only access information in their own language, or have low literacy.

Motivated by this, we investigate using three high resource Indian language ASR systems (Hindi, Telugu, and Tamil) for STD of ten low-resource Indian languages, most of which currently do not have any ASR systems available. Since there is no standard dataset for STD in these languages, we create our own dataset by using a publicly available TTS dataset in these languages. We first experiment with not using any language-specific data at all, and then add a text Language Model some target languages. We also bootstrap the lexicons for all target languages as g2p systems are not available for them. Combining the outputs of the three ASR systems and using language-specific text data performs best, without the need for language-specific speech data.

\section{Relation to Prior Work}

Both Query-by-Example and lexical STD have been well studied in the low resource setting. \cite{anguera2014query} describes the results of the MediaEval 2013 challenge where the task was QbE for 9 low resource languages with a diverse channel and environment conditions. \cite{mendels2015improving} scrapes text data from the web to improve low-resource keyword spotting. \cite{ni2016cross} describes an approach for data selection for cross-lingual transfer learning that leads to large improvements in STD performance. \cite{soto2014rescoring} re-scores confusion networks for low resource STD using various features such as lexical, phonetic, false alarm and structural features. \cite{ma2014strategies} extract features such as word burstiness, confusion network posteriors, acoustic and prosodic features to improve STD for low resource languages.

\cite{li2014empirical} use a multilingual DNN-based ASR with a shared hidden layer and cross-lingual model transfer for STD in 7 low resource languages. \cite{sun2017empirical} use transfer learning for cross-lingual STD from a high-resource language English to a low resource language German. The IARPA BABEL \cite{gales2014speech} and DARPA LORELAI \cite{christianson2018overview} programs have led to a lot of research in low-resource STD. \cite{knill2013investigation} investigate language-independent multilingual models for low-resource STD and find that adaptation to the target language is required for reasonable performance.

In this work, we focus on STD of 10 low resource Indian languages, most of which do not have any ASR systems available. We use three higher resource Indian language ASR systems and evaluate their effectiveness alone or combined in a multilingual setting for STD. In contrast to some of the approaches mentioned earlier, we use the ASR systems as a black-box and only utilize the transcriptions provided by the ASR systems to search for target words. All the languages we consider are Indian languages, and we discuss the implications of phonetic relatedness among these languages in the results. 


 \section{Data and Resources}

\begin{table}[!h]
\caption{TTS Dataset statistics \cite{baby2016resources}}
\label{tab:datastats}
{\renewcommand{\arraystretch}{\pad}
   \begin{tabular}{?>{\columncolor[HTML]{EFEFEF}}l|l|l?}
 \hbl
  \cellcolor[HTML]{EFEFEF}\textbf{Language}    & \cellcolor[HTML]{EFEFEF}\textbf{Hrs}  & \cellcolor[HTML]{EFEFEF}\textbf{Utts}      \\ \hline
   Assamese   & 51    & 35k       \\ \hline
   Bengali      & 30    & 18k        \\ \hline
   Gujarati   & 41    & 15k       \\ \hline
   Kannada      & 34    & 20k       \\ \hline
   Malayalam  & 35    & 22k       \\
 
  \hbl
   \end{tabular}
\quad
   \begin{tabular}{?>{\columncolor[HTML]{EFEFEF}}l|l|l?}
  \hbl
   \cellcolor[HTML]{EFEFEF}\textbf{Language}    & \cellcolor[HTML]{EFEFEF}\textbf{Hrs}  & \cellcolor[HTML]{EFEFEF}\textbf{Utts}      \\ \hline
  Marathi     & 8   & 4k        \\ \hline
   Odia          & 8.5   & 7k        \\ \hline
   Bodo      & 4      &2.7k       \\ \hline
   Manipuri     & 41    & 38k       \\ \hline
   Rajasthani & 33    & 18k       \\
  \hbl
   \end{tabular}

   }

 \end{table}

 We need the following data and resources to build an STD system: terms in the target low resource languages along with the audio to be searched, lexicons, ASR systems in high resource languages and text data to build Language Models (LMs) in the target languages if available. We describe all these resources in detail in this section.
 
 \subsection{Terms and Audio}

 We use data recorded for Speech Synthesis in Indian languages available at \footnote{https://www.iitm.ac.in/donlab/tts/voices.php} \cite{baby2016resources}. This dataset contains audio from 13 Indian languages. We use audio from Assamese, Bengali, Gujarati, Kannada, Malayalam, Marathi, Odia, Bodo, Manipuri and Rajashthani. Table \ref{tab:datastats} contains the number of hours and utterances available for each language. This data consists of clear speech recordings from professional speakers and corresponding transcripts in the native script.

We select 1000 utterances from each target language to use as the test audio for our task. We also randomly select 100 keywords of length 1-10 characters, 50 keywords of length 10-15 characters and 10 keywords of length 15-20 characters as the terms that we will search for from the 1000 utterances in each language. 

 \subsection{Lexicon}

 Since we use systems cross-lingually, it is important that all the languages share the same phoneset. The Hindi, Telugu and Tamil ASRs share a common phoneset, so we need a Grapheme to Phoneme (g2p) mapping for all the other target languages in terms of this phoneset. One way to do this is to manually map all characters in the target languages to the ASR phoneset, however, since we need to map ten target languages to Hindi, Tamil and Telugu, we automate the process as much as possible.
 
 First, we use the Unitran mapping \cite{qian2010python} available as part of the Festvox Indic frontend \cite{parlikar2016festvox} to obtain a mapping between all Unicode characters and a symbol from the X-SAMPA phoneset for all target languages as well as the languages of our ASR systems. Then, we obtain the phonemes for each Unicode character in Hindi, Tamil and Telugu using the ASR system's g2p system. This gives us a mapping between the X-SAMPA phonemes from Unitran and the ASR system phonemes for all three languages. We create a dictionary of X-SAMPA - ASR phonemes and verify that the mapping between the phonemes is the same across languages. In cases where the mapping does not hold, we remove the dictionary item.

 Next, we generate a list of all X-SAMPA phonemes for all the target languages by looking up the Unitran table for all the Unicode characters present in each language. We use the dictionary created earlier to map the X-SAMPA phonemes for each langauge to an ASR phoneme. A few phonemes in each language remain unmapped, because the phonemes do not exist in Hindi, Telugu or Tamil. For these cases, we manually map the leftover phonemes to the closest ones in Hindi. We use Hindi as a pivot language to obtain mappings from Telugu and Tamil to all the target languages and manually map any leftover phonemes to the closest phonemes in Telugu and Tamil. 
 

 One important point to note here is that we do not use a g2p system that uses context for any of the ten languages, which is a limitation of this approach, particularly for languages which have many contextual rules such as schwa deletion, voicing rules and contextual nasalization.

\subsection{ASR systems}

 We use state-of-the-art ASR systems trained on hundreds of hours of data in Hindi, Tamil and Telugu for our experiments. The Hindi ASR is trained on significantly more data compared to Tamil and Telugu and is expected to perform better. We run all three ASRs on the test data and calculate the Phone Error Rate (PER) as shown in Table \ref{tab:PER}. The Hindi ASR has a PER of ~25\% on Gujarati test data while Tamil ASR has a PER of 73\% on the same data, which is expected given that Gujarati is phonetically similar to Hindi and very different from Tamil. Similarly, we find that the Telugu ASR achieves the lowest PER on Kannada, while the Tamil ASR achieves the lowest PER on Malayalam. Overall, we obtain lowest PERs of 20-25\% on Marathi and Gujarati which are very similar to Hindi, while the highest PER is ~80\% for Bodo which is a Sino-Tibetan language and is most different from all three ASR languages.

 \begin{table}[!h]
\caption{PER of the Hindi, Telugu and Tamil ASRs for the target languages shown in table \ref{tab:datastats}}
   \label{tab:PER}
 {\renewcommand {\arraystretch}{\pad}
 \resizebox{0.45\textwidth}{!}{
   \begin{tabular}{?>{\columncolor[HTML]{EFEFEF}}l|l|l|l?}
   \hbl
     \cellcolor[HTML]{EFEFEF}Languages & \cellcolor[HTML]{EFEFEF}Hi-In   & \cellcolor[HTML]{EFEFEF}Te-In  & \cellcolor[HTML]{EFEFEF}Ta-In   \\
              \cellcolor[HTML]{EFEFEF}&\cellcolor[HTML]{EFEFEF} (\%)  \cellcolor[HTML]{EFEFEF} &\cellcolor[HTML]{EFEFEF} (\%)  \cellcolor[HTML]{EFEFEF}&  \cellcolor[HTML]{EFEFEF}(\%)   \\\hline
   Assamese  & \textbf{69} & 79 & 88   \\\hline
   Bengali   & \textbf{58} & 68    & 80  \\ \hline
   Bodo      & \textbf{82} &  94   & \textbf{82}  \\\hline
   Gujarati      & \textbf{25} & 54    & 73  \\\hline
   Kannada      & 60 & \textbf{35}    & 69 \\
\hbl

   \end{tabular}
   \quad
   \begin{tabular}{?>{\columncolor[HTML]{EFEFEF}}l|l|l|l?}
\hbl
   \cellcolor[HTML]{EFEFEF}Languages &\cellcolor[HTML]{EFEFEF} Hi-In   & \cellcolor[HTML]{EFEFEF}Te-In  & \cellcolor[HTML]{EFEFEF}Ta-In   \\
             & \cellcolor[HTML]{EFEFEF}(\%)   & \cellcolor[HTML]{EFEFEF}(\%)  & \cellcolor[HTML]{EFEFEF}(\%)   \\\hline
   Malayalam  & 64 & 59    & \textbf{50}   \\ \hline
   Manipuri   & \textbf{64} & \textbf{64}    & 80  \\ \hline
   Marathi      & \textbf{20} & 65    & 72\\ \hline
   Odia      & \textbf{59}  & 62    &79  \\ \hline
   Rajasthani & \textbf{34} & 66     & 78 \\ 
\hbl
   \end{tabular}
  
 } 
 }
   
 \end{table}

 \subsection{Language Models}

 Although speech data is scarce in the target languages that we consider, text data in the form of Wikipedia, news websites and user generated content may be available. We download Wikipedia dumps\footnote{https://dumps.wikimedia.org/$<$lang-code$>$/wiki/latest/} for Assamese, Bengali, Marathi, Kannada and Gujarati. We pre-process the data to remove all the English characters, special symbols, leading spaces and blank lines and use the text to build n-gram LMs. The perlexities (PPLs) on held-out test sets of Assamese and Kannada (around 900) are significantly higher than the other three languages (around 500) as shown in Table \ref{tab:LM_stats}. The high PPL values for these languages can be attributed to some extent to spelling variations.

 \begin{table}[!h]
\caption{N-gram Language Modeling perplexity on target language text data (Number of Sentences)}
  \label{tab:LM_stats}
 \renewcommand{\arraystretch}{\pad}
      \centering
  \begin{tabular}{?>{\columncolor[HTML]{EFEFEF}}l|l|l?}
   \hbl
\cellcolor[HTML]{EFEFEF}Languages & \cellcolor[HTML]{EFEFEF}Training Data  &   \cellcolor[HTML]{EFEFEF}n-gram PPL  \\ \hline
  Assamese&70k    & 962        \\\hline
  Bengali&70k      & 455         \\\hline
  Marathi&70k   & 439     \\\hline
  Kannada&70k   &   753     \\\hline
  Gujarati&70k   &  528      \\
\hbl
\end{tabular}
 \end{table}

 \begin{table}[!h]
\caption{Grouping of Phonemes to overcome spelling variations and spelling mistakes}
  \label{tab:phoneme_group}
 \renewcommand{\arraystretch}{\pad}
      \centering
  \begin{tabular}{?>{\columncolor[HTML]{EFEFEF}}l|l?}
  \hbl
     \cellcolor[HTML]{EFEFEF}Languages & \cellcolor[HTML]{EFEFEF}Training Data    \\  \hline
  Class A& `ih',`iy'      \\\hline
  Class B&  `uh', `uw', `ow', `oy'             \\\hline
  Class C&  `ay', `ae'     \\\hline
  Class D& `dt', `dr'     \\\hline
  Class E& `q', `k'     \\\hline
  Class F& `qh', `kh'     \\\hline
  Class G& `gd', `g'     \\\hline
  Class H& `jh', `jhh'     \\\hline
  Class I& `dht', `dhh'     \\\hline
  Class J& `ng',`ny', `nd'     \\
  \hbl
  \end{tabular}
 
  
 \end{table}

 \section{Experiments}
 \label{sec:exps}

We decode all the 1000 utterances in each target language using all three ASR systems and search the resulting hypothesis text for keywords. We use an utterance level string match algorithm similar to the one proposed in \cite{shah2019}. We use a string alignment algorithm \cite{needleman1970general} to search for the query keyword in the ASR transcripts after converting both the transcripts and the keywords to the same phone set. Needleman-Wunsch \cite{needleman1970general} is a dynamic programming based algorithm and was originally used for aligning gene sequences. It takes two strings as parameters and outputs the alignment score between the two strings. We assign +2 for match, -1 for mismatch, -1 for gap and -0.1 for extended gap. If the alignment score is above a certain threshold we consider the keyword to be found. The scores and the threshold are determined empirically by us. Since we match phonemes in the target keyword and decoded hypotheses, we also create phonetic classes by grouping similar phonemes together and allow similar phonemes to be matched as shown in Table \ref{tab:phoneme_group}. 



\subsection{Baselines}
We use the following two methods as our baseline approaches:

\textbf{Cross-lingual}: In this approach we pass the test audio through the three ASRs - Hindi, Telugu, Tamil and search for the keyword is each of the transcripts respectively. We report the Mean Term Weighted Value scores for each language in Table \ref{tab:baselines}. 

\textbf{Multilingual}: In the second baseline approach we calculate the alignment score of the keyword with each of the three ASR transcripts and look for the keyword in the the transcript which has the maximum alignment score with the keyword. We report the Mean Term Weighted Value score considering the best match for each keyword.

\subsection{Language-specific LMs}

In this experiment, we replace the LM and lexicon of the high resource ASR systems with the target language's LM and lexicon. We use the Hindi ASR to decode Assamese, Bengali, and Gujarati, since they are all part of the Indo-Aryan language family and the Telugu ASR to decode Kannada, since they are both Dravidian languages. We pass each word in the LMs through Unitran and use the mapping between X-SAMPA and the Hindi and Telugu ASR phonemes to get lexicons in each language in terms of the ASR phonemes. Then, we decode as before and search for the terms in the new ASR transcripts.




\section{Evaluation and Results}
\label{sec:results}

\subsection{Evaluation metric}
\label{sec:EvalMetric}

We use a standard metric for Spoken Term Detection, Mean Term Weighted Value (MTWV) proposed for the NIST Spoken Term Detection evaluation \cite{fiscus2006spoken}, which is one minus the average value lost by the system per term. The value lost by the system is a weighted linear combination of Miss Probability $(P_{Miss})$ and false alarm probability $(P_{FA})$. MTWV of 1 indicates no misses and no false alarms, while a system that outputs nothing gets a MTWV of 0. Negative MTWVs are possible.


\begin{equation}
P_{Miss}(term,\theta) = 1 - \big(\frac{N_{C}(term,\theta)}{N_{T}(term)}\big)
\end{equation}

\begin{equation}
\vspace{-0.2cm}
P_{FA}(term,\theta) = \big( \frac{N_{spurious}(term,\theta)}{N_{NT}(term)}\big)  
\end{equation}

\vspace{-0.2cm}
\begin{equation}
AE(term,\theta) = \frac{P_{Miss}(term,\theta) +\beta *P_{FA}(term,\theta) }{N_{TT}} 
\end{equation}

\vspace{-0.2cm}
\begin{equation}
TWV(term,\theta) = 1 - AE(term,\theta)
\end{equation}

\vspace{-0.2cm}
\begin{equation}
MTWV(term,\theta) = AVG(TWV(term,\theta))
\end{equation}

The $\theta$ parameter is the detection threshold, $N_{C}(term,\theta)$ is number of correct (true) detections of the term with a detection score greater than or equal to $\theta$, $N_{spurious}(term,\theta)$ is the number of spurious (incorrect) detections of the term with a detection score greater than or equal to $\theta$, $N_{T}(term)$ is the true number of occurrences of the term in the corpus, $N_{NT}(term)$ is the number of opportunities for incorrect detections of the term in the corpus \cite{fiscus2006spoken}.

Depending on the application, different weights are assigned to $P_{Miss}$ and $P_{FA}$. The $\beta$ in eq. 3 is the weight parameter. For our case, we give equal weights to the $P_{Miss}$ and $P_{FA}$ by having the $\beta$ value equal to 1.

\subsection{Results}

Table \ref{tab:baselines} shows the MTWV scores for all ten languages decoded with the baseline approach, that is without the use of a language specific LM or lexicon. The Hindi ASR performs best on Assamese, Bengali, Bodo, Gujarati, Manipuri, Marathi and Rajasthani, while the Telugu ASR performs best on Kannada and Odia. The Tamil ASR performs best on Malayalam. This is similar to the trends observed in Table \ref{tab:PER} except in the case of Manipuri, where the PER of Telugu and Hindi are the same while the MTWV score is significantly lower for Telugu. Although the PER of Bodo for all three ASRs is very low, the MTWV is similar to the other languages. The example below shows the target keywords, original transcript and phonetic sequence of the Bodo audio, followed by the decoded transcript of the same audio by the Hindi ASR and corresponding phonetic sequence.
\\ 

 \begin{figure}[!htb]

    \center{\includegraphics[height=35mm,width=80mm]{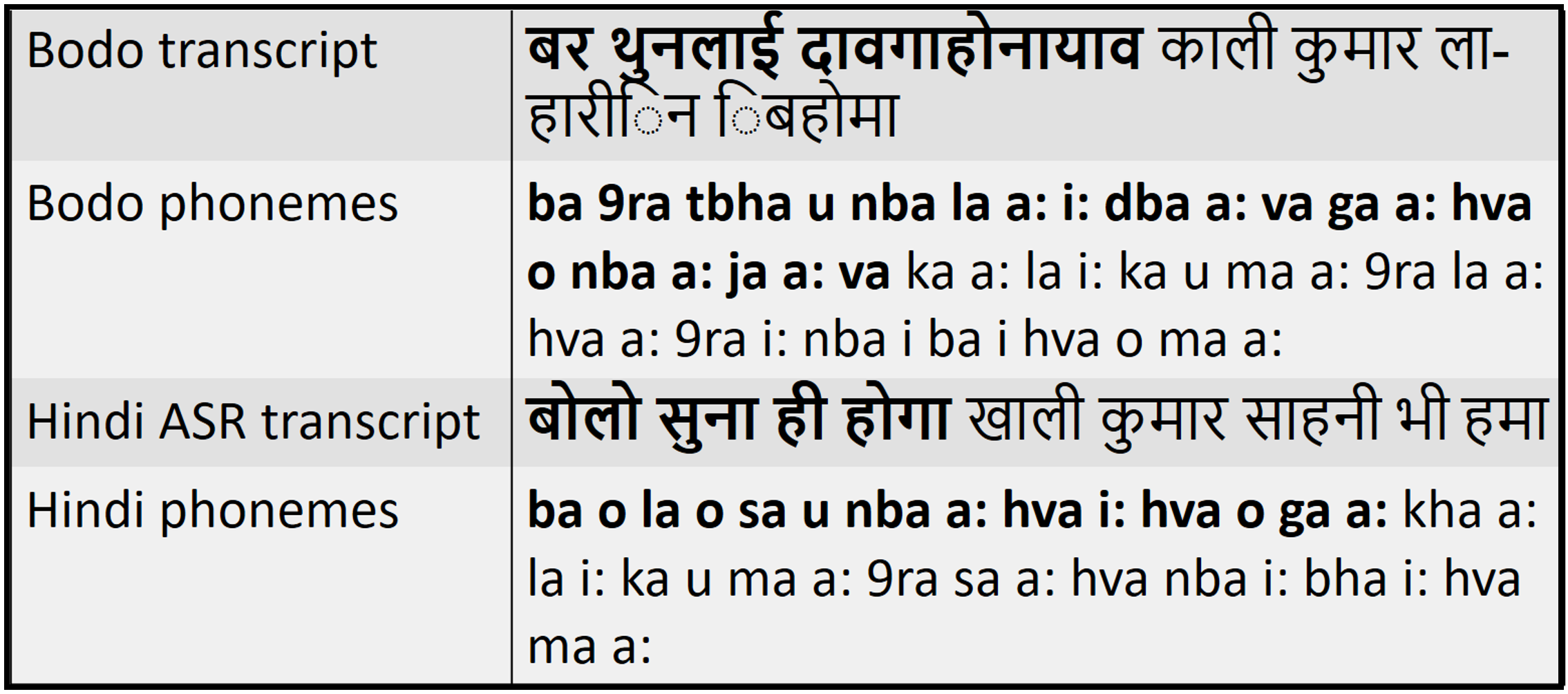}}
    \caption{Bodo keywords and its corresponding phonemes, Hindi ASR transcript and its corresponding phonemes are in bold}

\end{figure}





As we can see, even though there are differences in phonemes between the target words and decoded output, we manage to recover the keyword due to our matching algorithm that takes into account phonetic classes while matching, and allows similar phonemes to be matched. This leads to a high MTWV score for Bodo even though the PER is high. Overall, other than Bodo, we get the highest MTWV scores for Gujarati, Marathi and Rajasthani which are phonetically very similar to Hindi.

Next, we look at the combined results when the ASR hypotheses from all three languages is taken and the hypotheses that gives the highest alignment score is considered for matching. We find that for all languages except Gujarati in which the score remains unchanged, using all the transcripts leads to a higher MTWV score than considering a single language.

\begin{table}[!h]
 \caption{MTWV scores for the Baseline Method}
   \label{tab:baselines}
\renewcommand{\arraystretch}{\pad}
     \centering
   
  \begin{tabular}{?>{\columncolor[HTML]{EFEFEF}}l|l|l|l|l?}
  \hbl
 & \multicolumn{4}{c?}{\cellcolor[HTML]{EFEFEF}Baseline} \\\cline{2-5}
  \multirow{-2}{*}{Languages}& Hi-In   & Te-In  & Ta-In  & Multilingual  \\\hline
    
  Assamese  & \textbf{0.37} & 0.30   &   0.22    &   \textbf{0.39}\\\hline
  Bengali   & \textbf{0.46} & 0.27   &   0.23    &   \textbf{0.47}\\\hline
  Bodo      & \textbf{0.54} & 0.50   &   0.45    &   \textbf{0.56}\\\hline
  Gujarati  & \textbf{0.62} & 0.39   &   0.27    &   \textbf{0.62}\\\hline
  Kannada   & 0.35 & \textbf{0.36}   &   0.31    &   \textbf{0.41}\\\hline
  

  Malayalam & 0.28 & 0.25  &    \textbf{0.33}    &\textbf{0.37}\\\hline
  Manipuri  & \textbf{0.41} & 0.24  &    0.21    &\textbf{0.44}\\\hline
  Marathi   & \textbf{0.57} & 0.31  &    0.20    &\textbf{0.58}\\\hline
  Odia      & 0.39 & \textbf{0.40}  &    0.22    &\textbf{0.43}\\\hline
  Rajasthani& \textbf{0.52} & 0.41  &    0.29    &\textbf{0.55}\\
 \hbl
  \end{tabular}

\end{table}

Table \ref{tab:final_results} shows the MTWV values for Assamese, Bengali, Marathi and Gujarati decoded with the Hindi ASR and Kannada decoded with the Telugu ASR with language-specific LMs. We find that adding a language-specific LM leads to a significant improvement in MTWV scores in all cases.

 \begin{table}[!h]
\caption{MTWV scores of the Multilingual baseline system vs. system with language-specific LM}
  \label{tab:final_results}
 \renewcommand{\arraystretch}{\pad}
      \centering
  
  \begin{tabular}{?>{\columncolor[HTML]{EFEFEF}}l|l|l?}
  \hbl
  \cellcolor[HTML]{EFEFEF}Languages& \cellcolor[HTML]{EFEFEF}Baseline  & \cellcolor[HTML]{EFEFEF}+LM        \\  \hline
 
  Assamese& 0.39    & \textbf{0.49}        \\\hline
  Bengali&0.47      &  \textbf{0.79}      \\\hline
  Marathi&0.58   &  \textbf{0.75}    \\\hline
  Kannada&0.41   &    \textbf{0.86}    \\\hline
  Gujarati&0.62   &   \textbf{0.81}     \\
\hbl
  \end{tabular}
  
 \end{table}
 
\begin{figure}[!htb]

    \center{\includegraphics[height=35mm,width=90mm]{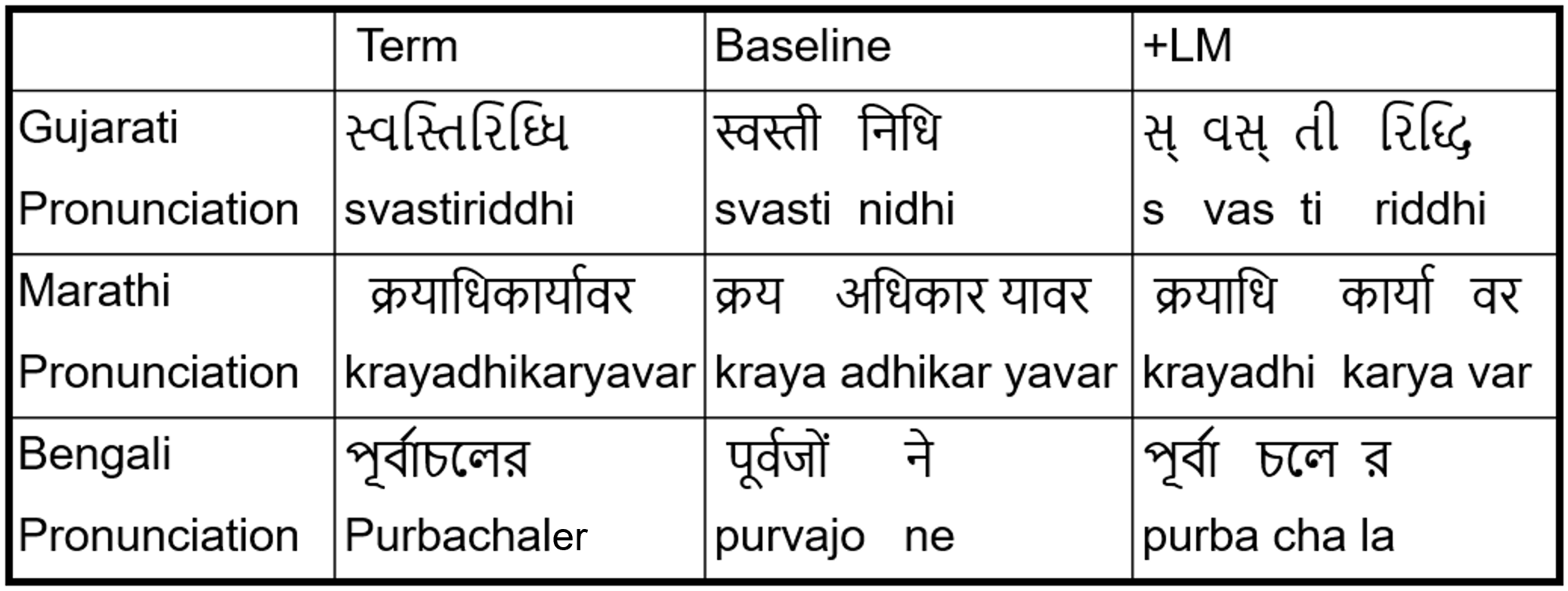}}
    \caption{Examples of the STD by the baseline model v/s baseline+LM using the Hindi decoder\label{fig:examples}}

\end{figure}

 Figure \ref{fig:examples} shows examples of terms in Gujarati, Bengali and Marathi decoded by the Hindi decoder with and without a language specific LM. As we can see, even though the baseline technique can find words that are phonetically similar to the target in the Hindi transcript, adding a language-specific LM increases the chances of finding the exact words or word fragments we are looking for. Note however that we still need to perform a string alignment-based match since the exact terms we are looking for are not present in the transcript.
 
\section{Conclusion}

In this work, we use three high-resource ASR systems for Spoken Term Detection in ten low-resource Indian languages, without the need for language-specific speech data. We show that it is possible to detect terms in languages without any language-specific LMs, particularly in the case where the target language is similar to the language of the decoder. However, the addition of a language-specific LM helps and may be easy to build given the availability of text data in the language.

We bootstrap lexicons for all the target languages using a resource that maps all Unicode characters to a single phone set. We also show that it is possible to perform STD on languages that are phonetically dissimilar to the target language (Bodo and Hindi) by using an algorithm that allows for phonemes from the same phonetic class to match, which results in a high MTWV score despite having a high Phone Error Rate. 

In future work, we would like to incorporate language specific information into decoding by rescoring the lattice, and also combine lattices from multiple ASR systems, or use a ROVER-like \cite{fiscus1997post} technique to come up with the best transcript from multiple recognition systems. The experiments conducted in this work were on clean TTS data which was recorded in a noise-free environment by professional speakers. In future work, we would like to develop robust techniques for found data, which could provide challenges such as background noise and music and lead to more ASR errors. Since no standard dataset for STD exists in these low-resource Indian languages, we plan to make the dataset we derived from the TTS dataset available for research use.

\newpage

\bibliographystyle{IEEEtran}
\bibliography{refs}

\newpage

 \end{document}